\pdfoutput=1
\documentclass{article}

\usepackage[nonatbib, preprint]{neurips_2021}

\usepackage[utf8]{inputenc} 
\usepackage[T1]{fontenc}    
\usepackage{hyperref}       
\usepackage{url}            
\usepackage{booktabs}       
\usepackage{amsfonts}       
\usepackage{nicefrac}       
\usepackage{microtype}      
\usepackage{xcolor}         
\usepackage{bm}
\usepackage{amsmath}
\usepackage{graphicx}

\title{ContourRender: Detecting Arbitrary Contour Shape For Instance Segmentation In One Pass}

\author{%
  Tutian Tang$^{1*}$\hspace{1em}
  Wenqiang Xu$^{1}$\thanks{These two authors contributed equally to this work.}\hspace{1em}
  Ruolin Ye$^1$\hspace{1em}
  Yan-Feng Wang$^1$\hspace{1em}
  Cewu Lu$^1$\hspace{1em}
  \\
  $^1$Shanghai Jiao Tong University\\
  \texttt{\{tttang,vinjohn,cathyye2000,wangyanfeng,lucewu\}@sjtu.edu.cn} \\
}

\begin{document}

\maketitle

\begin{abstract}
  Direct contour regression for instance segmentation is a challenging task. Previous works usually achieve it by learning to progressively refine the contour prediction or adopting a shape representation with limited expressiveness. In this work, we argue that the difficulty in regressing the contour points in one pass is mainly due to the ambiguity when discretizing a smooth contour into a polygon. To address the ambiguity, we propose a novel differentiable rendering-based approach named \textbf{ContourRender}. During training, it first predicts a contour generated by an invertible shape signature, and then optimizes the contour with the more stable silhouette by converting it to a contour mesh and rendering the mesh to a 2D map.
  This method significantly improves the quality of contour without iterations or cascaded refinements. Moreover, as optimization is not needed during inference, the inference speed will not be influenced.
  Experiments show the proposed ContourRender outperforms all the contour-based instance segmentation approaches on COCO, while stays competitive with the iteration-based state-of-the-art on Cityscapes. In addition, we specifically select a subset from COCO val2017 named COCO ContourHard-val to further demonstrate the contour quality improvements. Codes, models, and dataset split will be released.
\end{abstract}

\section{Introduction}
\begin{figure}
\centering
\includegraphics[width=\linewidth]{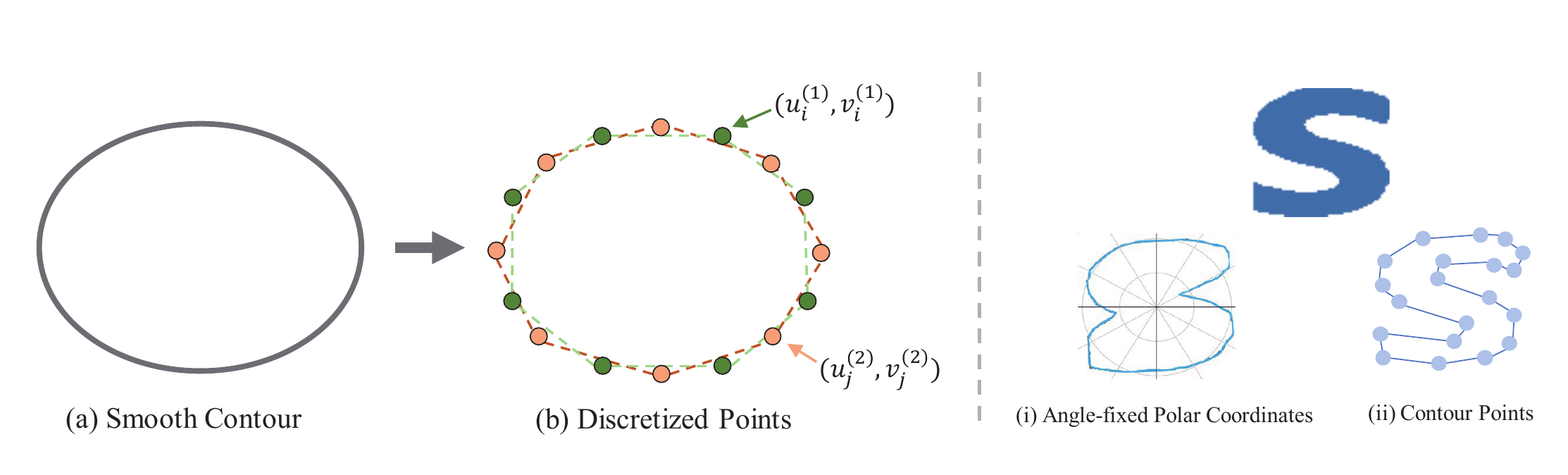}
\caption{The ambiguity of the contour point coordinates. \textbf{Left:} A continuous circle can be discretized in different ways. The inner areas of the discretized contours are similar but the coordinates are very different; \textbf{Right:} Angle-fixed polar coordinate representation (\cite{ese_seg,polarmask,star_medicine_contour}) is rather stable but compromise the expressiveness on concave shapes.}
\label{fig:coord_ambiguity}
\end{figure}

Instance segmentation is one of the fundamental tasks in computer vision. Aside from object detection, it also needs to predict the shape of the interested objects, which are usually represented by masks \cite{maskrcnn,mask_score_rcnn,panet,shape_mask,HTC_ins_seg,solo2} or contours \cite{ese_seg,polarmask,deepsnake,polygonrnn,polygonrnn++}. While the former is widely adopted by the mainstream researches, the contour-based approaches draw increasing attentions, mostly due to the advantage on the control of the shape which is beneficial to applications such as automatic annotation \cite{polygonrnn,polygonrnn++,curve_gcn}, autonomous driving \cite{levelset_1,levelset_2,deepsnake}, and medical image analysis \cite{active_contour_differential_render,star_medicine_contour}. Since a contour shape (polygon) is a closed curve which consists of several ordered points, predicting the contour shape is essentially regressing the point coordinates. As the coordinates can have large numeric range and are intrinsically unstable (See Fig. \ref{fig:coord_ambiguity}), previous works usually regress them cumbersomely during inference with iterative optimizations \cite{deepsnake,polygonrnn,polygonrnn++,active_contour_2,active_contour_3,active_contour_4,active_contour_differential_render}, cascaded optimizations \cite{contour_proposal}, or compromised representations \cite{ese_seg,star_medicine_contour,polarmask}. In this work, we explore to regress a decent contour in one pass without limitation on the shape representation.

Based on the observation from Fig. \ref{fig:coord_ambiguity}, it is natural to think that optimizing the inner content is more stable in comparison to directly regressing the point coordinates. 
To achieve this, we introduce two designs namely \textit{coordinate signature} and \textit{contour mesh}. A coordinate signature is transformed from point coordinates and introduced to facilitate the learning of coordinate information, thus it should be compact, robust, and invertible. Discrete cosine transform (DCT) \cite{dct} coefficients are adopted as the coordinate signature. Discussion about DCT coefficients and other choices of the coordinate signatures can be found in Sec. \ref{sec:dct_coef}. However, since the coordinate signature is only an interpretation of the coordinates and cannot address the problem of discretization ambiguity, we further convert the reconstructed coordinates to a contour mesh and render the contour mesh to a silhouette with an off-the-shelf differentiable renderer \cite{nmr}. The scheme of contour mesh construction is discussed in Sec. \ref{sec:contour_mesh}.
Since the differentiable renderer plays the central role, we name the proposed novel contour regression pipeline as \textbf{ContourRender}.
To show that the proposed ContourRender is independent of the object detector design, we equip it with the popular one-stage detector FCOS \cite{fcos} and two-stage Mask R-CNN \cite{maskrcnn}. The modification details are described in Sec. \ref{sec:struct_contour_render}.

To quantitatively evaluate the proposed ContourRender, we conducted the experiments on the challenging COCO \cite{coco} and Cityscapes \cite{cityscapes} datasets. To show the advantages over predicting complex contours, we select 358 images according to the shape convexity \cite{kins_amodal} from \textit{COCO val2017} datasets to form a \textit{COCO HardContour-val} subset. Our approach outperforms all the contour-based methods. Since the differentiable renderer is not used in the inference phase, the computational overhead to produce the contour is mainly originated from the coordinate signature prediction and inverse transformation, which is negligible since it can be easily parallelized.

We summarize our contributions in two folds. First, we propose a novel contour-based approach for instance segmentation named ContourRender. The ContourRender can achieve accurate contour prediction by DCT coordinate signature and differentiable contour mesh rendering without the need for iterative regression or compromise on the contour representation. Second, the ContourRender adapted from Mask R-CNN or FCOS both outperform previous contour-based baselines on COCO, Cityscapes, and our proposed challenging COCO HardContour-val subset.

\section{Related Works}\label{sec:related_work}
\paragraph{Contour-based Instance Segmentation}
For many applications, contour representation is of particular interest due to the ability to explicitly describe the shape with points. Though we can certainly convert the object masks produced by the mask-based instance segmentation frameworks \cite{maskrcnn,mask_score_rcnn,panet,shape_mask,HTC_ins_seg,solo2} to contours with post-processing \cite{opencv}, it is still worthy to explore an end-to-end scheme to obtain the contour. Such attempts can be roughly divided into three categories based on their inference behaviors: one-pass approaches, iterative approaches, and cascaded approaches.

\textit{One-pass approach.} Since the coordinate representation is hard to be regressed directly, ESE-Seg \cite{ese_seg} adopted a polar-coordinate representation and approximate the $(r, \theta)$ function with Chebyshev polynomials. Later, PolarMask \cite{polarmask} took a similar design, with a polar IOU which directly optimizes the contour IOU. The polar-coordinate system enables sampling the contour point along the fixed angle sequence, which can largely reduce the ambiguity of the Cartesian coordinate system (from 2 DoFs to 1 DoF). However, it also limits the expressive ability on the concave shapes (Fig. \ref{fig:coord_ambiguity}). Another attempt is to regress from a pre-defined point set anchor, as in \cite{pointsetnet}. However, according to Appendix \ref{sec:appendix_offset}, the statistics about the offset ranges from the pre-defined point set anchor and the origin point have no significant difference.

\textit{Iterative approach.} A well-known iterative approach is active contour \cite{snakes}. Since \cite{active_contour_1} incorporated the active contour into an end-to-end learning framework, many improvements have been proposed \cite{active_contour_2,active_contour_3,active_contour_4,deepsnake,active_contour_differential_render}. Among them, \cite{active_contour_differential_render} integrated active contour with a differentiable renderer. The ideology is similar to ours, but their approach requires iterations and the authors only apply it to rather simple shapes like buildings or road scenes. DeepSnake \cite{deepsnake} proposed a circle convolution operator to adapt to the polygon representation. It achieves state-of-the-art performance over contour-based instance segmentation frameworks on several benchmarks. Aside from the active contour-based methods, by regarding the contour with different perspectives, RNNs (regarded as ordered point sequence) \cite{polygonrnn,polygonrnn++} and GCN (regarded as point graph) \cite{curve_gcn} are also adopted to address the polygon regression problem. They all require iteration operations to achieve satisfactory results.

\textit{Cascaded approach.} Unlike the iterative approaches that refine the prediction progressively with the shared modules, the cascaded approach usually designs individual components for prediction refinement. In \cite{contour_proposal}, after the contour proposals are generated, the coarse prediction will be improved by a local refinement process where the refined offsets are learned individually.

\paragraph{Differentiable Renderer} The rendering process can be viewed as mapping from a mesh to an image. Though many factors can be taken into consideration when making the rendering differentiable, such as gradient approximation \cite{opendr,nmr,mesh_gradient_1,mesh_gradient_2}, rasterization \cite{mesh_rendering_1,mesh_rendering_2,mesh_rendering_3} and ray-tracing \cite{mesh_illum_1,mesh_illum_2,mesh_illum_3,mesh_illum_4,mesh_illum_5}. As silhouette is all we need, rasterization and ray-tracing-based neural renderers are rather complicated. Thus we adopt a simple renderer \cite{nmr}. It obtains the differentiable gradient by sampling in a smooth manner. On the other hand, point cloud-based neural renderers \cite{render_point_1,render_point_pytorch3d,render_point_2} seem more straightforward than mesh with respect to the contour points. However, in practice, we find the point cloud renderer cannot bring out the improvement effect as the mesh renderers do. We have two speculations on this observation: (1). when the point radius is too large, the coordinate information is blurred in the circles, which introduces another kind of coordinate ambiguity. (2). when the point radius is too small, the coordinate information is predominant, as well as the coordinate ambiguity mentioned in Fig. \ref{fig:coord_ambiguity}. This motivates us to design the contour mesh (Sec. \ref{sec:contour_mesh}).

\begin{figure}
\centering
\includegraphics[width=\linewidth]{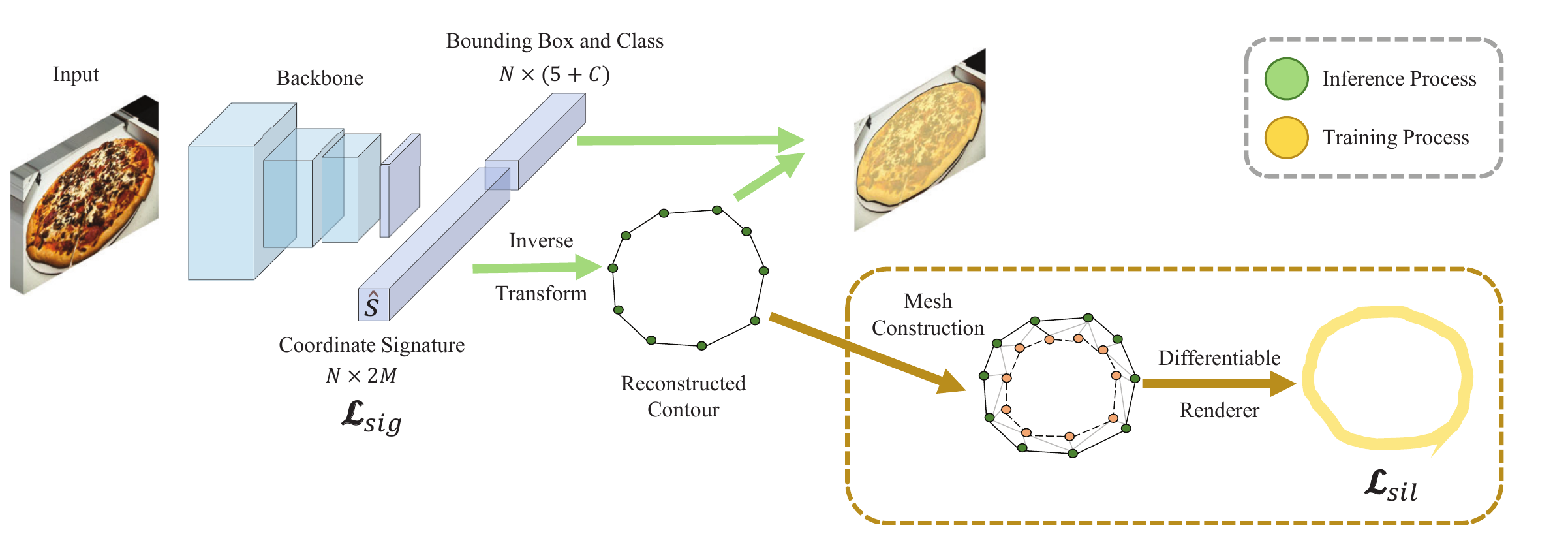}
\caption{The overall pipeline of ContourRender. The feature map will be feedforwarded to an MLP to obtain the signature coefficients $\hat{S}$, and then an inverse transformation is applied on $\hat{S}$ to obtain the predicted contour points $\bm{p'}$. Following, the contour points are integrated into the contour mesh, and the mesh will be put through the differentiable renderer to obtain a silhouette $\hat{M}$. All the operations in the ContourRender branch are fully differentiable so that it can be learned end-to-end. And the contour mesh construction and the differentiable rendering process can be removed in the inference phase, thus the inference time is not affected.}
\label{fig:pipeline}
\end{figure}

\section{Method}
Given an image $\mathcal{I} \in \mathbb{R}^{W\times H\times 3}$, a general contour-based instance segmentation framework aims to predict all the locations $\mathcal{O}=\{\bm{o}_i=(x_i, y_i, w_i, h_i)\}_{i=1}^N$ and the contour shapes $\mathcal{P}=\{\bm{p}=\{u_{ij}, v_{ij}\}_{j=1}^M\}_{i=1}^N$, where $N$ is the number of the detected objects, and $M$ is the number of the points for each contour shape.

For contour regression, we first predict the discrete cosine transform (DCT) coefficients of the contour shape and obtain a coarse contour with the inverse DCT (iDCT) operation (Sec. \ref{sec:dct_coef}). We convert the coarse contour to a contour mesh and render the corresponding silhouette after that(Sec. \ref{sec:contour_mesh}).

For the object detection part, we suggest two choices namely Mask R-CNN \cite{maskrcnn} and FCOS \cite{fcos} and demonstrate how to modify them to adapt with the ContourRender in Sec. \ref{sec:object_det}. To note, the ContourRender can be similarly extended to other object detectors.

If not specified, we set $M = 32$ for main analysis and experiments. How point number affects the performance please refer to Appendix \ref{sec:appendix_point_num}.
The overall pipeline is illustrated in Fig. \ref{fig:pipeline}.

\subsection{Coordinate Signature}\label{sec:dct_coef}
As mentioned earlier, the coordinate signature should be compact, robust, and invertible to fit into the neural network learning paradigm. There are many candidates, such as Cartesian coordinates, polar coordinates, dictionary coefficients of the Cartesian coordinates, DCT descriptor of the Cartesian coordinates, etc.

In the main paper, we will compare the Cartesian coordinates and their dictionary and DCT descriptors to illustrate how we decide the most preferable one before training. The experiments reported in Sec. \ref{sec:ablative} confirms the foresight. For the other signatures, please refer to Appendix \ref{sec:appendix_signatures}.

\subsubsection{Computation of the Signatures and the Invertibility} \label{sec:invertible}

Given a contour shape $\bm{p}=\{(u_j, v_j)\}_{j=1}^M$, the computation of the signatures and the inverse transform to reconstructed contour shape $\bm{p'}$ are described as the following.

\textbf{Coordinate Signature} 
Given a ground-truth bounding box $(x_c, y_c, w, h)$, we have $\bm{p}_c = (x_c,  y_c)^T$, $\bm{p}_s = (w, h)^T$, and $S_{coord}=\frac{\bm{p}-\bm{p}_c}{\bm{p}_s} \in \mathbb{R}^{M\times 2}$,
where $\bm{p}_c$ stands for the center and $\bm{p}_s$ stands for the scale. Inversely, during inference, given the predicted bounding box, the exact position and scale can be obtained by $S_{coord}\times \hat{\bm{p}_c} + \hat{\bm{p}_s}$. Note that all operations here are \textit{element-wise}.

\textbf{Dictionary Coordinate Signature} $S_{dict}$ are given by optimizing $||S_{coord} - S_{dict}D||^2_2 + ||S_{dict}||_1$ with Orthogonal Matching Pursuit (OMP) \cite{omp} which follows \cite{dict_learn}, where $D$ is the learned dictionary atoms. To recover $\bm{p'}$, it first converts $S_{dict}$ back to $S_{coord}$ by $S_{coord}=S_{dict} \times D$, and then convert to $\bm{p'}$ following the same procedure as coordinate signature.

\textbf{DCT Coordinate Signature} $S_{dct} = DCT(S_{coord})$, the applied DCT is 1-D for each axis \cite{dct}. Similarly, to obtain $\bm{p'}$, it first converts $S_{coord} = iDCT(S_{dct})$, and then convert to $\bm{p'}$ in the same way with coordinate signature.

It is noteworthy some common choices \cite{contour_survey} of contour signatures are not invertible, such as centroid distance function and contour curvature.

\subsubsection{The Representation Power and Compactness} 
\begin{figure}
\centering
\includegraphics[width=0.9\linewidth]{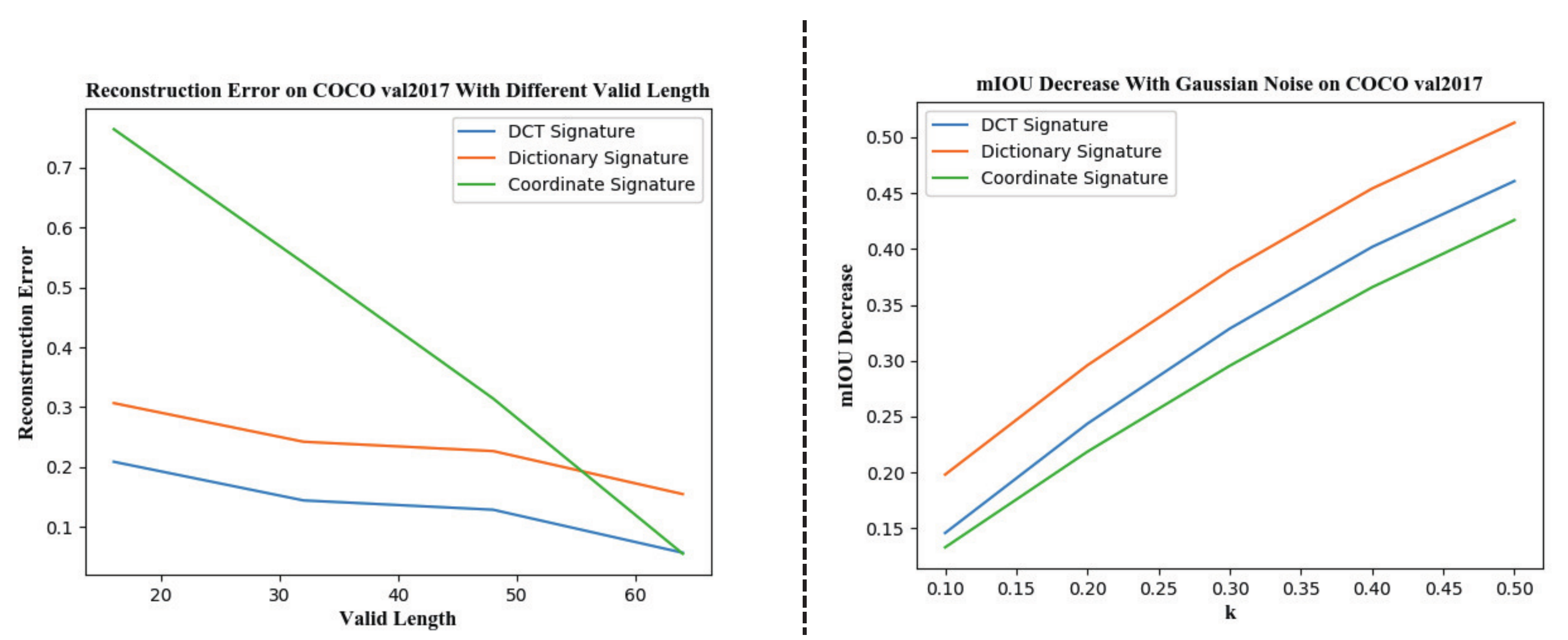}
\caption{\textbf{Left:} Reconstruction Error on COCO val2017 w.r.t valid length; \textbf{Right:} mIOU on COCO val2017 affected by different variance of Gaussian noise.}
\label{fig:compact_and_robust}
\end{figure}

Compactness means that the signatures should be able to aggregate most information to a few coefficients with a constrained numeric range. Empirically, we find that a more compact signature is easier to learn. Unfortunately, we do not have rigid theory for this observation, but some discussions are made in Appendix \ref{sec:appendix_compact}.
We conduct an offline analysis on the COCO val2017 dataset, which consists of 36781 instance shapes of 80 categories. We evaluate three signatures on the reconstruction ability concerning the valid coefficient length (number of non-zero coefficients), as shown in Fig. \ref{fig:compact_and_robust}. The reconstruction error is defined as $1-\text{mIOU}$. $IOU$ is calculated by comparing the ground truth mask and the reconstructed mask by filling the contour polygon and resizing to the original size. When reducing the valid coefficient length, we gradually set the coefficients to zero from the last elements to the first.

\subsubsection{Sensitivity Analysis and Robustness}
As the neural networks are known to regress with noise, it is informed to test the sensitivity of the signatures. We randomly add Gaussian noise $\mathcal{N}\sim (0, k\times e_i)$ to each coefficient element $e_i$ in the signatures. $k$ is set to 0.1, 0.2, 0.3, 0.4, 0.5. We show how much mIOU loss will the Gaussian noises cause on different signatures. The results are shown in Fig. \ref{fig:compact_and_robust}.

\subsubsection{On the Choice of the Coordinate Signature} We have compared three coordinate signatures from three perspectives. Among them, all the signatures can be invertible to reconstruct the original contour shape. While coordinate signature has the best performance on resisting the noise, it is not compact enough. Thus the DCT signature is adopted. Admittedly, it is possible there exist better coordinate signatures, we left it as the future works.

\subsubsection{Signature Loss}
Given the signature $S$ obtained from the ground truth mask, we can directly supervise the signature with:
\begin{equation}
    \mathcal{L}_{sig} = SmoothL1(\hat{S}, S),
\end{equation}
where the $\hat{S} \in \mathbb{R}^{M\times 2}$ is the predicted signature. $SmoothL1(\cdot)$ is the smooth L1 loss.

After learning on the signature loss, we can already obtain a coarse contour, by converting the predicted $\hat{S}$ according to the procedures described in Sec. \ref{sec:invertible}. But as mentioned earlier, signature learning cannot address the ambiguous nature of the coordinates. Thus we propose to train on a more stable silhouette through differentiable rendering.

\subsection{Contour Mesh}\label{sec:contour_mesh}
According to the discussion in Sec. \ref{sec:related_work} and the results from Appendix \ref{sec:appendix_dr}, we adopt NMR \cite{nmr} as the off-the-shelf renderer. In this case, we should convert the contour into a contour-derived mesh (short for contour mesh) before rendering.

\subsubsection{Construction of Contour Mesh}
As the differentiable gradient from the neural renderer will back-propagate to the vertex, i.e. the contour points, the construction process should try to avoid damaging contour points.

A mesh is composed of 3D vertex and triangle faces. 3D vertex can be simply obtained by stacking all-zero z-axis to the contour points, while the ways of constructing the faces need to be discussed.

\begin{figure}
\centering
\includegraphics[width=0.9\linewidth]{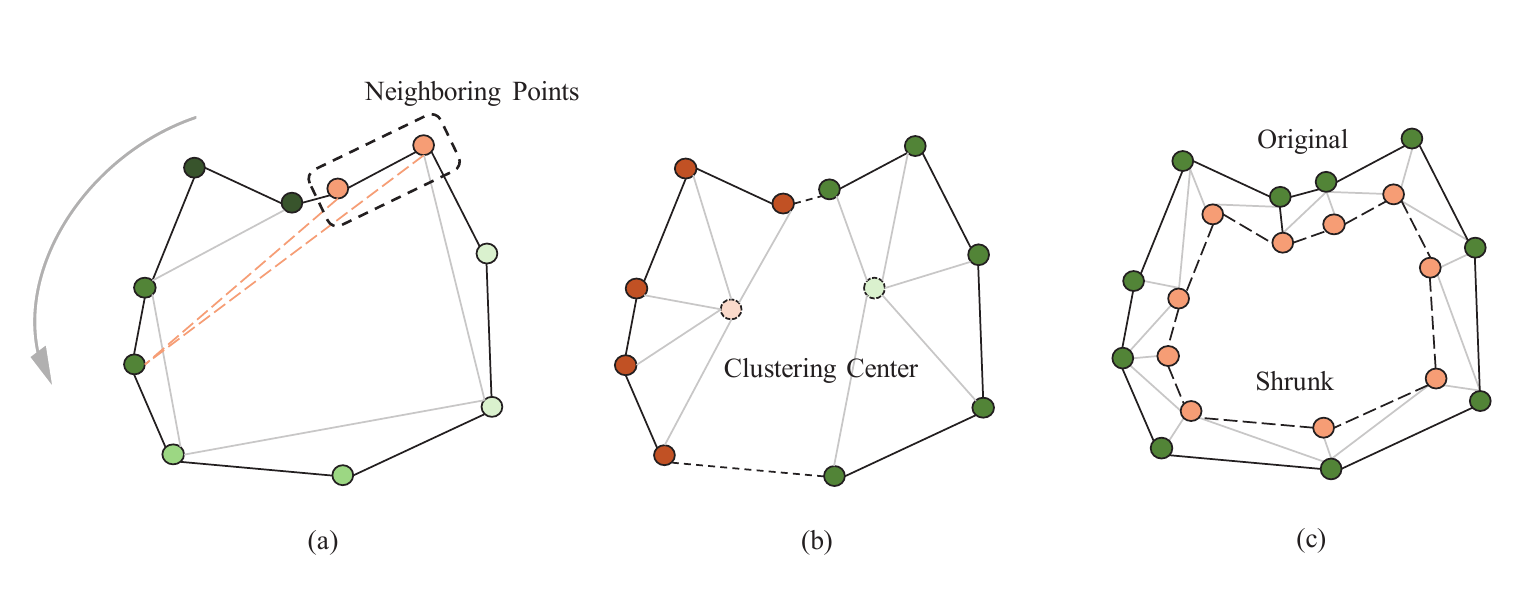}
\caption{Different intuitive ways to construct different kinds of contour meshes: (a) Internal Linking; (b) External Linking - KMeans; (c) External Linking - Shrink.}
\label{fig:contour_mesh}
\end{figure}

There are two intuitive approaches to construct the contour mesh, i.e. \textbf{Internal Linking} and \textbf{External Linking}. Internal linking means constructing the mesh faces based on the existing contour points, while external linking means we first find a set of the external points and link the contour points with them to form the faces.
Both approaches can be conducted with different strategies. Here we give some intuitive thoughts and select one for the main experiments. The illustration of linking strategies is displayed in Fig. \ref{fig:contour_mesh}. 

\textit{Internal Linking.} For internal linking, each neighboring point pair will forward counter-clockwise to find the first point which can make these three points a triangle with an area larger than a threshold $t$. We set $t=0.1$ for all the datasets.  

\textit{External Linking.} For external linking, an instant idea is to perform the clustering algorithm such as KMeans \cite{kmeans} to obtain the clustering centers. The clustering centers are most likely to be inside the contour. And then we can link the contour points to their corresponding clustering centers to form a contour mesh. This process is denoted as 'External Linking-KMeans'. However, as we would like to have the linking process run on the fly, in practice, we take another simple and faster approach. As the contour predicted is already zero-centered (because the supervised signature is obtained after zero-centered), we can simply shrink the contour by multiplying a scaling factor $s$ less than 1, and link the corresponding points between original and shrunk contours to form the contour mesh. We set $s=0$ for all the datasets. Different choices of the scaling factor are discussed in Appendix \ref{sec:appendix_scaling_factors}. This process is named as 'External Linking-Shrink'.

\subsubsection{Silhouette Loss}
After we transform the contour points to contour mesh, we can generate the corresponding silhouette by NMR \cite{nmr}. The silhouette is denoted as $\hat{M}$, while we put the ground truth contour forward the same process and result in $M$. In this way, the silhouette loss is defined by:
\begin{equation}
    \mathcal{L}_{sil} = \mathcal{F}(\hat{M}, M),
\end{equation}
$\mathcal{F}(\cdot)$ is a mask-based loss. In practice, we adopt Lovasz Softmax \cite{lovasz}, but we also discuss the common choices of MSE loss, BCE loss, and Dice loss in Sec. \ref{sec:ablative}. 

\subsection{ContourRender with Object Detectors}\label{sec:object_det}
\begin{figure}
\centering
\includegraphics[width=\linewidth]{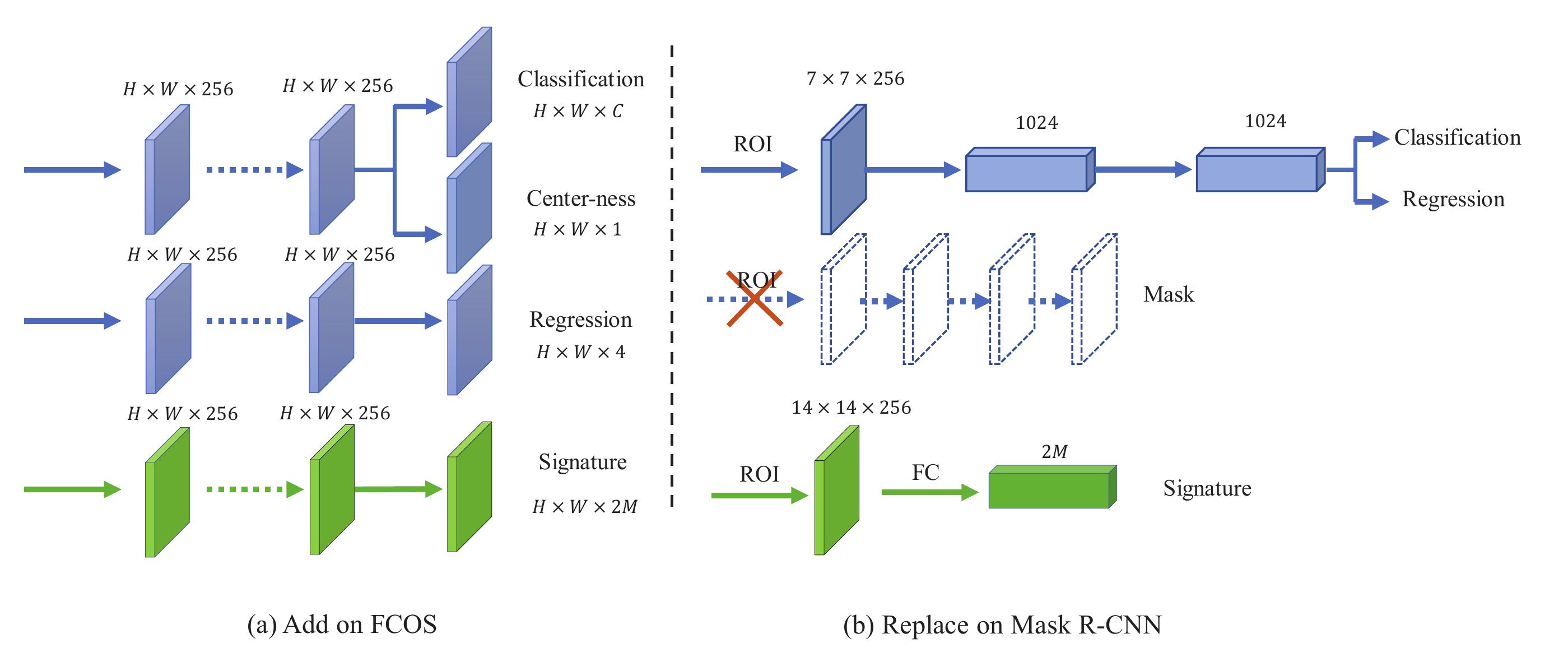}
\caption{Head Modification on One-Stage detector FCOS and Two-stage Mask R-CNN.}
\label{fig:structure_modification}
\end{figure}
\subsubsection{ContourRender Branch}\label{sec:struct_contour_render}
A generic ContourRender branch is designed as shown in Fig. \ref{fig:pipeline}. In this section, we will describe how it can be integrated into existing object detectors. For those who can only predict the object location like FCOS, we show how to add the ContourRender branch to enable shape prediction. For those object detection frameworks that already can do instance segmentation like Mask R-CNN, we show how to replace the original mask branch with our ContourRender branch. The modifications are shown in Fig. \ref{fig:structure_modification}.

\textbf{Add Modification of FCOS}
As the FCOS can only perform object detections, we \textit{add} the ContourRender branch by taking the same input feature with the box head. 

\textbf{Replace Modification of Mask R-CNN}
As the Mask R-CNN is already an instance segmentation framework, we \textit{replace} the original mask branch to the proposed ContourRender branch. The original mask features with the shape of $14\times 14$ after the RoiAlign operators are directly flattened and fed into the ContourRender branch. 

\subsubsection{Overall Objective Functions}
With the object detector, we denote the object detection objective functions as $\mathcal{L}_{det}$, the details of $\mathcal{L}_{det}$ for each detector remain the same, and can be referred to the original papers \cite{maskrcnn, fcos}. Therefore, the overall learning objective function for ContourRender is given as follows:
\begin{equation}
    \mathcal{L}_{all} = \mathcal{L}_{det} + \lambda_1\mathcal{L}_{sig} + \lambda_2\mathcal{L}_{sil},
\end{equation}
where $\lambda_1$ and $\lambda_2$ are the adjusting weights for different loss, which are determined by cross-validation and discussed in Sec. \ref{sec:impl}.

\section{Experiments}
\subsection{Implementation Details}\label{sec:impl}

We implement our method in two public frameworks, e.g. MMDetection \cite{mmdetection} and Detectron2 \cite{wu2019detectron2}.
The default backbone is ResNet-50 with ImageNet-pretrained weights.
We train the network for 150K iterations with stochastic gradient descent (SGD), which finished within 18 hours on 4 RTX Titan GPUs.
The initial learning rate is 0.02 and the mini-batch size is 16. The learning rate is reduced to 0.001 and 0.0001 at 60k and 140k, respectively. Weight decay and momentum are set to 0.0001 and 0.9, respectively. To stabilize the training process, we set $\lambda_1=1$ and $\lambda_2 =0$ during the first 60K iterations. Then, we set $\lambda_2=1$ till the end and make $\lambda_1$ decay linearly to $0$ till 70K. The performance gap between the two frameworks is negligible.

\subsection{Datasets}
In this work, we will use three datasets to evaluate the performance of the ContourRender on Mask R-CNN and FCOS, namely COCO \cite{coco}, Cityscapes \cite{cityscapes} and COCO ContourHard-val.
\paragraph{COCO} COCO is a large-scale dataset with 860001 shapes for training and 36781 shapes for validation. It has large shape varieties including a fair amount of separated shapes. To deal with the separated shapes, we just connect the separate contours by stacking them to form a larger contour. 

\paragraph{COCO ContourHard-val} As most COCO object shapes are fairly round, like shapes in bear, stop sign, and frisbee, etc., even imperfect contour representation can obtain fairly good performance. Thus, to better compare the contour representation, we select a subset from COCO val2017 as the COCO ContourHard-val set. The selection is based on the convexity mentioned in \cite{kins_amodal}. Only images with object shape of convexity smaller than 0.4 are selected to be validated. As a result, 358 images are chosen, which contains 5054 shapes of 77 categories. Statistics for each category please refer to Appendix \ref{sec:appendix_contourhard}.

\paragraph{Cityscapes} Cityscapes is a popular road-scene dataset with 2975 images 54060 shapes for training, and 500 images with 10415 shapes for validation. It includes 8 categories mostly concerning people and vehicles on the road.

\subsection{Results on COCO}
\paragraph{Comparison with Other Baselines}
We compare our methods with both mask and contour-based methods, it shows our method is competitive with mask-based methods, especially, our methods only drops 1 AP by replacing the mask branch of Mask R-CNN to ContourRender. Considering contour is rather limited to express separated shapes, or hollowed shapes, a little performance loss is expected.

With regarding to contour-based methods, for one-pass approach baseline PolarMask \cite{polarmask}, in the same setting with FCOS \cite{fcos} as the base detector, our method outperform 1.5 AP on COCO val2017, and 2.6 AP on COCO ContourHard-val (Table \ref{tab:coco_hard_results}). Besides, our methods also beat the iterative approach DeepSnake \cite{deepsnake}, even though the latter approach use the iterative optimization to refine the contour while ours do not.

\begin{table}[]
\centering
\begin{tabular}{c|c|c|c|c|c|c|c}
    Method & Backbone & AP & AP$_{50}$ & AP$_{75}$ & AP$_S$ & AP$_M$ & AP$_L$  \\\hline
    \textit{Mask-based}\\
    Mask R-CNN \cite{maskrcnn} & Res50FPN & 33.6 & 55.2 & 35.3 & - & - & -\\
    Mask R-CNN \cite{maskrcnn} & Res101FPN & 35.7 & 58.0 & 37.8 & 15.5 & 38.1 & 52.4\\
    PANet \cite{panet} & Res50FPN & 36.6 & 58.0 & 39.3 & 16.3 & 38.1 & 53.1 \\
    SOLOv2 \cite{solo2} & Res50FPN & 38.8 & 59.9 & 41.7 & 16.5 & 41.7 & 56.2\\
    \hline
    \textit{Contour-based} \\
    DeepSnake \cite{deepsnake} & DLA-34 & 30.3 & - & - & - & - & - \\
    ESE-Seg \cite{ese_seg} & DarkNet-53 & 21.6 & 48.7 & 22.4 & - & - & - \\
    PointSetNet \cite{pointsetnet} & RX101FPN-DCN & 34.6 & 60.1 & 34.9 & - & - & - \\
    PolarMask \cite{polarmask} & Res50FPN & 29.1 & 49.5 & 29.7 & 12.6 & 31.8 & 42.3 \\
    PolarMask \cite{polarmask} & Res101FPN & 30.4 & 51.9 & 31.0 & 13.4 & 32.4 & 42.8 \\
    PolarMask \cite{polarmask} & RX101FPN-DCN & 36.2 & 59.4 & 37.7 & 17.8 & 37.7 & 51.5 \\
    \hline
    Ours (F) & Res50FPN & 30.6 & 51.6 & 30.8 & 14.7 & 31.5 & 40.3\\
    Ours (F) & Res101FPN & 31.7 & 53.4 & 32.2 & 15.6 & 33.8 & 45.4\\
    Ours (F) & RX101FPN-DCN & 37.9 & 61.1 & 38.2 & 20.3 & 39.6 & 52.7\\
    Ours (M) & Res50FPN & 32.6 & 54.6 & 33.8 & 15.9 & 34.6 & 46.7 \\
    Ours (M) & Res101FPN & 34.8 & 56.8 & 36.4 & 18.1 & 36.8 & 49.2\\
    Ours (M) & RX101FPN-DCN & 41.0 & 62.8 & 42.5 & 23.9 & 43.4 & 54.1 \\\hline
\end{tabular}
\caption{AP metrics on COCO val2017. Backbone: Res50FPN means ResNet-50 \cite{resnet} with FPN structure \cite{fpn}, Res101FPN can be inferred; RX means ResNext \cite{resnext}; DCN is the deformable convolution \cite{dcn}; DarkNet is introduced in \cite{yolov3}; DLA is introduced in \cite{deepsnake}. Ours (F) means base detector is FCOS, and ours (M) means it is adapted from Mask R-CNN.}
\label{tab:coco_results}
\end{table}

\paragraph{Comparison with Other Baselines On COCO ContourHard-val}
Since the validation dataset is built specifically for contour comparison, we only compare the contour-based methods on this validation set. As shown in Table \ref{tab:coco_hard_results}, our method beats the baseline methods by a large margin.

\begin{table}
\parbox{.45\linewidth}{
    \centering
    \begin{tabular}{c|c|c|c|c}
        Method & Backbone & AP & AP$_{50}$ & AP$_{75}$  \\\hline
        ESE-Seg* & Res50FPN & 22.3 & 41.6 & 22.7\\
        PolarMask & Res50FPN & 24.3 & 43.2 & 24.5 \\\hline
        Ours (F) & Res50FPN & 26.9 & 45.6 & 27.1\\
        Ours (M) & Res50FPN & 27.8 & 48.3 & 28.3\\\hline
    \end{tabular}
    \caption{AP metrics on COCO ContourHard-val. *: We change the base detector of ESE-Seg to FCOS to make fair comparison.}
    \label{tab:coco_hard_results}
}
\hfill
\parbox{.45\linewidth}{
\centering
    \begin{tabular}{c|c|c}
        Method & Backbone & AP   \\\hline
        PolygonRNN++ \cite{polygonrnn++} & Res50M & 22.8 \\
        DeepSnake* \cite{deepsnake} & DLA-34 & 28.2 \\
        DeepSnake \cite{deepsnake} & DLA-34 & 31.7 \\\hline
        Ours (M) & Res50FPN & 28.8 \\\hline
    \end{tabular}
    \caption{AP metrics on Cityscapes test split. *: DeepSnake w/o cascaded detection to handle fragmented instances, as well as ours. In this case, even though it inferences iteratively, ours is better. Res50M: A modified ResNet-50 structure.}
    \label{tab:city_ap}
}
\end{table}

\begin{figure}
\centering
\includegraphics[width=1\linewidth]{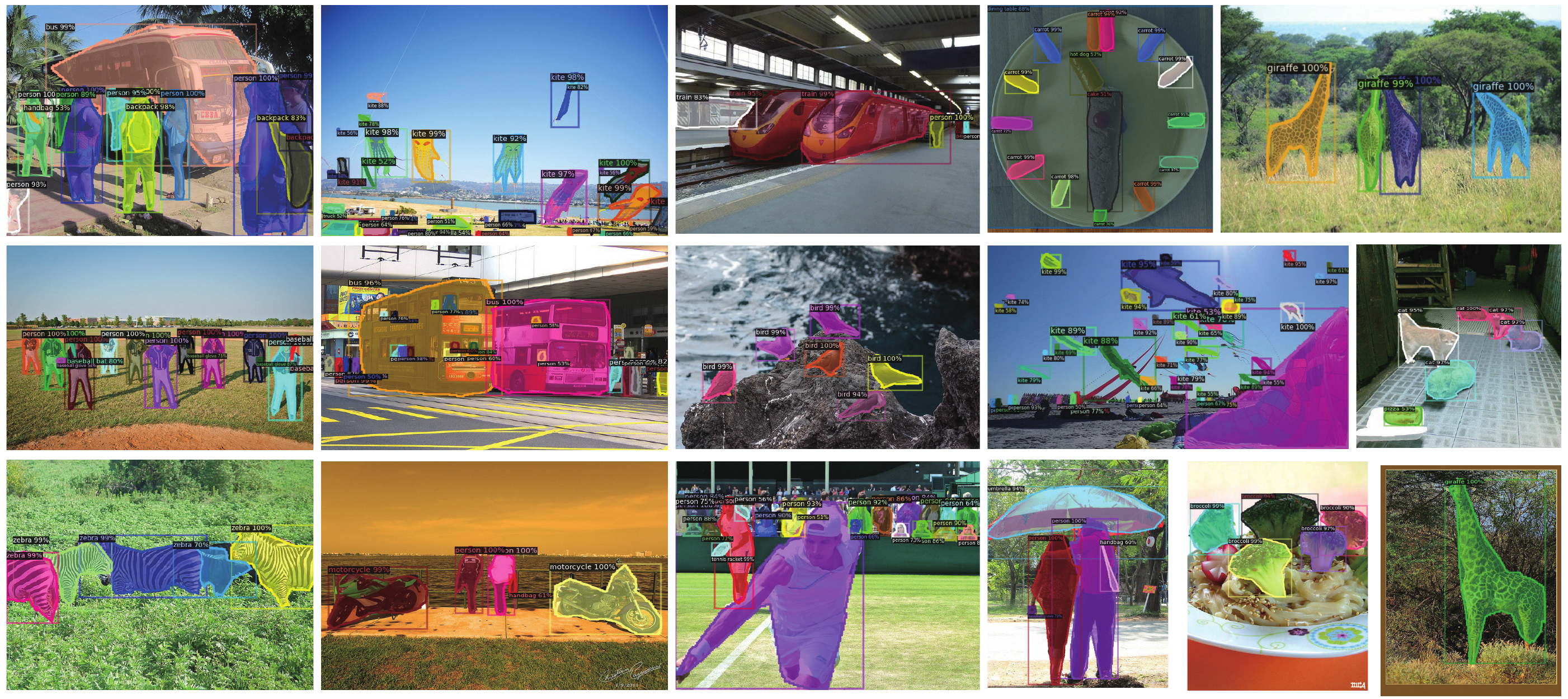}
\caption{The qualitative results on COCO test-dev.}
\label{fig:qual_coco}
\end{figure}

\subsection{Results on Cityscapes}
Since most contour-based approach report the mIOU metric on Cityscapes, we decide to follow the same training/testing and evaluation protocol. We notice that a few methods \cite{deepsnake,polygonrnn++} report AP metric on Cityscapes, to compare with them, we also report the AP metric on Cityscapes in Table \ref{tab:city_ap}. It is interesting to find that though almost all these contour-based methods are iterative approaches, our one-pass approach can outperform most of them.
\begin{table}[]
    \centering
    \begin{tabular}{c|c|c|c|c|c|c|c|c|c}
         Method & mIOU & person & rider & car & truck & bus & train & mcycle & bicycle \\\hline
         PolygonRNN \cite{polygonrnn} & 61.4 & 64.0 & 60.6 & 71.2 & 68.0 & 69.5 & 53.7 & 52.1 & 52.1 \\
         PolygonRNN++ \cite{polygonrnn++} & 70.2 & 70.8 & 68.5 & 78.0 & 77.9 & 79.6 & 62.8 & 61.7 & 62.3 \\
         SplineGCN \cite{curve_gcn} & 72.1 & 72.5 & 70.6 & 80.2 & 79.1 & 81.7 & 65.9 & 62 & 64.8 \\
         Gur et al. \cite{active_contour_differential_render} & \textbf{75.1} & \textbf{75.0} & \textbf{72.0} & \textbf{82.0} & \textbf{79.6} & 83.0 & \textbf{74.5} & \textbf{66.5} & \textbf{68.1} \\\hline
         Ours (M) & 73.5 & 73.4 & 69.5 & 79.8 & 79.1 & \textbf{83.6} & 73.1 & 64.4 & 65.2 \\\hline
    \end{tabular}
    \caption{Performance on Cityscape. Baseline methods are all iterative approaches, but our method can outperform most of them.}
    \label{tab:city_miou}
\end{table}

\subsection{Ablative Study}\label{sec:ablative}
The ablative studies are carried out on the COCO dataset with ContourRender-MaskRCNN (Res50FPN). More ablative studies can be referred to supplementary materials.

\textbf{The necessity of the signature learning} If we remove the signature learning process, that is, set $\lambda_1 = 0$ and $\lambda_2 = 1$ from beginning to end, we find that $\mathcal{L}_{sil}$ will not converge at all.

\textbf{Performance of different signatures} With the Cartesian coordinate signature, we obtain 24.5 AP; With the dictionary signature, we obtain 26.1 AP; With the DCT signature, we obtain 28.1 AP.

\textbf{Performance of different loss on the silhouette loss} Lovasz Softmax \cite{lovasz} is proved to be effective to serve as IOU loss, so we adopt it in ContourRender pipeline. If we substitute it to the common MSE loss, BCE loss and Dice loss, then the AP drops 4.1, 6.7, and 3.5 respectively.

\section{Limitations and Future Works}
In this work, we propose a novel contour-based instance segmentation pipeline, which enables the learning framework to produce decent contours without iterative or cascaded design, and the shape representation is not compromised as in previous works like ESE-Seg \cite{ese_seg} or PolarMask \cite{polarmask}. Nevertheless, our method still suffers from several drawbacks: 1. the contour representation cannot handle separated or hollowed contours at this moment. 2. the whole performance is largely dependent on the bounding box detection, which can give the information about the $\bm{\hat{p}}$ and the exact scale of the contour. 3. The choice of the coordinate signature and contour mesh is determined by intuition, which may be improved in later works.


\bibliographystyle{plain}
\bibliography{contour_render}

\section{Appendix}

Optionally include extra information (complete proofs, additional experiments and plots) in the appendix.
This section will often be part of the supplemental material.

\subsection{How Contour Point Number Affects Performance?}\label{sec:appendix_point_num}
\begin{figure}[ht!]
\centering
\includegraphics[width=1\linewidth]{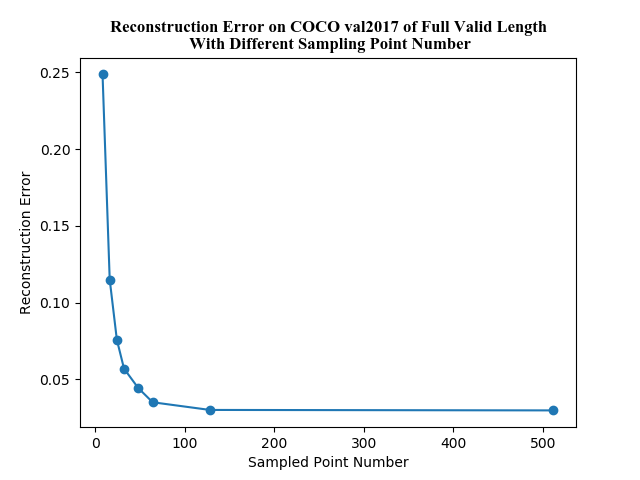}
\caption{The Reconstructed Error on COCO val2017 of full valid length with different sampled point numbers.}
\label{fig:reconstructed_error_number}
\end{figure}

Apparently, as the sampled point number grows, the reconstructed error will be consistently reduced, but since the contour cannot represent fragmented shapes for now, the lower bound of the reconstructed error on COCO val2017 is prone to 0.03. As shown in Fig. \ref{fig:reconstructed_error_number}.

However, the online experiments with ContourRender Mask R-CNN (Res50FPN) converges faster. As shown in Table \ref{tab:point_number}, when the sampled point number is larger than 32, the performance has no significant improvement. The shape signatures in these experiments are DCT signatures.
\begin{table}[h]
    \centering
    \begin{tabular}{c|c|c|c}
        \# Sampled Points & AP & AP\@50 & AP@75 \\\hline
        16 & 27.6 & 49.9 & 28.7\\
        24 & 30.4 & 52.6 & 31.2\\
        32 & 32.6 & 54.6 & 33.8\\
        40 & 32.7 & 55.1 & 33.2\\\hline
    \end{tabular}
    \caption{ContourRender Mask R-CNN (Res50FPN) performance on COCO val2017 with respect to different sampled point numbers.}
    \label{tab:point_number}
\end{table}

\subsection{Comparison with Other Signatures}\label{sec:appendix_signatures}
\begin{figure}[ht!]
\centering
\includegraphics[width=1\linewidth]{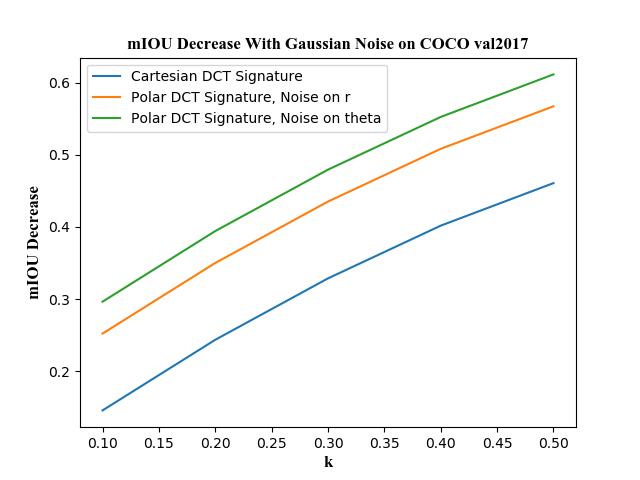}
\caption{mIOU decrease on COCO val2017 with different level of Gaussian noise on $r$ and $\theta$ respectively. In contrast, the mIOU decrease of Cartesian coordinate representation is also displayed.}
\label{fig:reconstructed_error_sig}
\end{figure}

In this section, we specifically compared with DCT signatures converted from polar coordinate systems. First, we think the angle-fixed sampling polar coordinate can largely damage the shape representation concerning convex shapes. Though it is not that significant in COCO dataset, it is not the natural representation after all. Thus, the polar coordinate representation is the $(r, \theta)$ coordinate with 2 DOFs. And we convert it to DCT signatures like Cartesian coordinates. However, we find such representation is quite sensitive to the noise, especially the $\theta$ element. As shown in Fig. \ref{fig:reconstructed_error_sig}.

\subsection{Why Compactness Matters?}\label{sec:appendix_compact}
The compactness in our context is similar to the concept \textit{sparsity}. But as some signatures do not have the nature of \textit{sparsity}, such as Cartesian coordinate signature, we determine to use a different concept. When we check the sparsity of the DCT and dictionary coefficients, it is surprisingly to find that even though we took a certain number $n$ valid coefficients, $n<2M$, the online experiments with neural network can maintain a fairly good performance. We believe it is a good property for a shape signature so that the neural network can learn to regress a small number of coefficients to achieve good results.

\subsection{Comparison with Other Differentiable Renderers}\label{sec:appendix_dr}
We have tested our pipeline with other differentiable renderers, such as \cite{render_point_2, render_point_pytorch3d, opendr,mesh_rendering_2}. 

For the point-based renderer \cite{render_point_pytorch3d}, during training, when $\lambda_2$ is set to 1, the validation score keeps dropping till 0. Similar phenomenon exhibits on another point-based renderer \cite{render_point_2}.

For the mesh-based renderer, we also tried with OpenDR \cite{opendr} and Soft Rasterizer \cite{mesh_rendering_2}. The behaviors are consistent with NMR \cite{nmr}. Since rendering a contour mesh towards a silhouette is a rather simple task.

\subsection{How Scaling Factor of External Linking - Shrink Affects the  Performance}\label{sec:appendix_scaling_factors}
As the External linking scheme is shrinking the contour, the resulting silhouettes would be $s\times 100\%$ full masks. $s$ the scaling factor and $0\leq s \leq 1$. When $s \to 0$, the resulting silhouette will be close to a full mask, while $s \to 1$, the resulting silhouette will be close to a thin contour. In practice, we find a large $s$ could cause the predicted silhouette and the ground truth silhouette hard to be matched even though the shape is similar to each other, but the IOU will still be very low. Thus, we compare different $s$ in Table \ref{tab:scaling_factor} and decide to simply set $s=0$. 

\begin{table}[h]
    \centering
    \begin{tabular}{c|c|c|c}
        s & AP & AP@50 & AP@75 \\\hline
        0 & 32.6 & 54.6 & 33.8\\
        0.3 & 32.8 & 54.5 & 33.9\\
        0.6 & 31.5 & 52.7 & 32.3\\
        0.9 & 30.1 & 50.3 & 30.9\\
        0.95 & 0 & 0 & 0\\\hline
    \end{tabular}
    \caption{ContourRender Mask R-CNN (Res50FPN) performance on COCO val2017 with respect to different scaling factors.}
    \label{tab:scaling_factor}
\end{table}

On the other hand, the shrinking scheme can trivially extended to dilating scheme, with a scaling factor $s > 1$.

\subsection{Do Offsets from Prior Templates Significantly Different From Coordinate Itself?} \label{sec:appendix_offset}
As previous works like active contour-based approaches \cite{} and other template-based \cite{pointsetnet} propose to regress a shape from a prior template. It maybe useful for iterative schemes, however, we find it is not that significantly different from the direct coordinate regression on COCO val2017. Given $M=32$ evenly sampled contour points, we compare the average offset (mean$_{\Delta x}$, mean$_{\Delta y}$), offset variance (var$_{\Delta x}$, var$_{\Delta y}$) on both x and y axis with different offset-schemes, like "offset to original points" (i.e. the coordinate it self), "offset from an outer box" \cite{pointsetnet}, "offset from a circle" \cite{active_contour_differential_render}. The online training with ContourRender Mask-RCNN (Res50FPN) also confirms the judgement, as shown in Table \ref{tab:offset}.
\begin{table}[h]
    \centering
    \begin{tabular}{c|cccc|ccc}
        Offset From & mean$_{\Delta x}$ & mean$_{\Delta y}$ & var$_{\Delta x}$ & var$_{\Delta y}$ & AP & AP@50 & AP@75 \\\hline
        Origin & -0.0017 & 0.0036 & 0.2139 & 0.2736 & 32.6 & 54.6 & 33.8 \\
        Outer Box & 0.0017 & 0.0036 & 0.2437 & 1.2098 & 31.8 & 53.3 & 32.6\\
        Circle & 0.0369 & 0.0071 & 0.7272 & 0.7801 & 31.7 & 53.4 & 32.6 \\\hline
    \end{tabular}
    \caption{Performance on COCO val2017 with respect to different offset schemes.}
    \label{tab:offset}
\end{table}

\subsection{Statistics of COCO ContourHard-val}\label{sec:appendix_contourhard}
In this section, we will give some statistics data about the selected COCO ContourHard-val set. As mentioned in the main paper, 358 images are chosen, which contains 5054 shapes of 77 categories. In table \ref{tab:contour_hand}, we list the number count and decrease (in brackets) from the original COCO val2017 for each category. 
\begin{table}[]
    \centering
    \begin{tabular}{c|c|c|c|c|c}
        category & \#instances & category & \#instances & category & \#instances  \\\hline
        person & 1500 \textbf{(-9277)} & bicycle & 34 \textbf{(-280)} & car & 193 \textbf{(-1725)} \\
        motorcycle & 23 \textbf{(-344)} &airplane & 27 \textbf{(-116)} & bus & 13 \textbf{(-270)} \\
        train & 11 \textbf{(-179)} & truck & 37 \textbf{(-377)} & boat & 95 \textbf{(-329)}\\
        traffic light & 28 \textbf{(-606)} & fire hydrant & 2 \textbf{(-99)} & stop sign & 1 \textbf{(-74)} \\
        parking meter & 7 \textbf{(-53)} & bench & 61 \textbf{(-350)} & bird & 143 \textbf{(-284)}\\
        cat & 9 \textbf{(-193)} & dog & 15 \textbf{(-203)} & horse & 29 \textbf{(-243)} \\
        sheep & 71 \textbf{(-283)} & cow & 32 \textbf{(-340)} & elephant & 8 \textbf{(-244)} \\
        bear & 0 \textbf{(-71)} & zebra & 15 \textbf{(-251)} & giraffe & 10 \textbf{(-222)} \\
        backpack & 90 \textbf{(-281)} & umbrella & 84 \textbf{(-323)} & handbag & 118 \textbf{(-422)} \\
        tie & 30 \textbf{(-222)} & suitcase & 34 \textbf{(-265)} & frisbee & 7 \textbf{(-108)} \\
        skis & 76 \textbf{(-165)} & snowboard & 14 \textbf{(-55)} & sports ball & 9 \textbf{(-251)} \\
        kite & 160 \textbf{(-167)} & baseball bat & 10 \textbf{(-135)} & baseball glove & 9 \textbf{(-139)} \\
        skateboard & 39 \textbf{(-140)} & surfboard & 28 \textbf{(-239)} & tennis racket & 17 \textbf{(-208)} \\
        bottle & 99 \textbf{(-914)} & wine glass & 47 \textbf{(-294)} & cup & 156 \textbf{(-739)} \\
        fork & 42 \textbf{(-173)} & knife & 64 \textbf{(-261)} & spoon & 56 \textbf{(-197)} \\
        bowl & 109 \textbf{(-514)} & banana & 20 \textbf{(-350)} & apple & 26 \textbf{(-210)}\\
        sandwich & 18 \textbf{(-159)} & orange & 39 \textbf{(-246)} & broccoli & 12 \textbf{(-300)} \\
        carrot & 39 \textbf{(-326)} & hot dog & 18 \textbf{(-107)} & pizza & 39 \textbf{(-245)} \\
        donut & 44 \textbf{(-284)} & cake & 24 \textbf{(-286)} & chair & 412 \textbf{(-1359)} \\
        couch & 27 \textbf{(-234)} & potted plant & 78 \textbf{(-264)} & bed & 5 \textbf{(-158)} \\
        dining table & 153 \textbf{(-542)} & toilet & 1 \textbf{(-178)} & tv & 25 \textbf{(-263)} \\
        laptop & 20 \textbf{(-211)} & mouse & 8 \textbf{(-98)} &remote & 29 \textbf{(-254)} \\
        keyboard & 13 \textbf{(-140)} & cell phone & 26 \textbf{(-236)} & microwave & 5 \textbf{(-50)} \\
        oven & 14 \textbf{(-129)} & toaster & 0 \textbf{(-9)} & sink & 13 \textbf{(-212)}\\
        refrigerator & 13 \textbf{(-113)} & book & 211 \textbf{(-918)} & clock & 16 \textbf{(-251)} \\
        vase & 30 \textbf{(-244)} & scissors & 6 \textbf{(-30)} & teddy bear & 2 \textbf{(-188)} \\
        hair drier & 0 \textbf{(-11)} & toothbrush & 1 \textbf{(-56)} & total & 5054 \textbf{(-31281)} \\
        \hline
    \end{tabular}
    \caption{Numbers of instances for categories in COCO ContourHard-val and Numbers of instances which are removed from the original COCO val2017 dataset due to too convex.}
    \label{tab:contour_hand}
\end{table}

\subsection{Broader Impact}\label{sec:boarder_impact}
\paragraph{Who may benefit from this research?}
Instance segmentation may benefit various applications, such as autonomous driving, robot manipulation/navigation. Especially when our proposed method can support real-time feedback as proper base object detector is adopted.

\paragraph{Who may be put at disadvantage from this research?}
We cannot think of a case when someone be put to disadvantage specifically because of instance segmentation methods. Maybe if a person on the sidewalk is mis-classified or miss-detected, he could be at risk from autonomous vehicle.

\paragraph{What are the consequences of failure of the system?}
There are three kinds of failures of the system, namely mis-classification, wrong localization, bad shape prediction.

\paragraph{Whether the task/method leverages biases in the data?}
We do not specifically leverages biases in the data, as the data and the metric is standard for this problem.

\end{document}